\pdfoutput=1

\documentclass[11pt]{article}

\usepackage[]{naacl2021}

\usepackage{times}
\usepackage{latexsym}

\usepackage[T1]{fontenc}

\usepackage[utf8]{inputenc}

\usepackage{graphicx}
\usepackage{microtype}
\usepackage{algorithm}
\usepackage[noend]{algpseudocode}
\usepackage{color,soul}
\usepackage{comment}
\usepackage{soul}
\usepackage[hide]{boxnotes}
\newcommand{\eat}[1]{}

\newcommand{\gan}[1]{\noindent{\textcolor{black}{\{{$_{gan}$} \bf   #1\}}}}
\iftrue 

  \renewcommand\gan[1]{\xspace}
\fi
\usepackage{array}
\newcolumntype{H}{>{\setbox0=\hbox\bgroup}c<{\egroup}@{}}
\usepackage{xcolor}

\title{Exploring Text-to-Text Transformers for English to Hinglish Machine Translation with Synthetic Code-Mixing}

\author{{\bf
  Ganesh Jawahar$^{1,2}$ \quad El Moatez Billah Nagoudi$^1$} \\
  {\bf Muhammad Abdul-Mageed$^{1,2}$  \quad  Laks V.S. Lakshmanan$^{2}$} \\
\normalsize Natural Language Processing Lab$^{1}$  \\
\normalsize Department of Computer Science$^{2}$  \\
  \normalsize The University of British Columbia\\
  \texttt{ \small ganeshjwhr@gmail.com, \{moatez.nagoudi, muhammad.mageed\}@ubc.ca, laks@cs.ubc.ca} 
}

\begin{document}
\maketitle
\begin{abstract}

We describe models focused at the understudied problem of translating between monolingual and code-mixed language pairs. More specifically, we offer a wide range of models that convert monolingual English text into Hinglish (code-mixed Hindi and English). Given the recent success of pretrained language models, we also test the utility of two recent Transformer-based encoder-decoder models (i.e., mT5 and mBART) on the task finding both to work well. Given the paucity of training data for code-mixing, we also propose a dependency-free method for generating code-mixed texts from bilingual distributed representations that we exploit for improving language model performance. In particular, armed with this additional data, we adopt a curriculum learning approach where we first finetune the language models on synthetic data then on gold code-mixed data. We find that, although simple, our synthetic code-mixing method is competitive with (and in some cases is even superior to) several standard methods (backtranslation, method based on equivalence constraint theory) under a diverse set of conditions. Our work shows that the mT5 model, finetuned following the curriculum learning procedure, achieves best translation performance ($12.67$ BLEU). Our models place first in the overall ranking of the English-Hinglish official shared task.

\end{abstract}
\section{Introduction}
\label{sec:intro}

\begin{table*}[h!]
\centering
\begin{tabular}{p{2.8in}p{0.1in}p{2.9in}} 
\hline 
\multicolumn{3}{c}{	\colorbox{red!10}{\textbf{Hinglish to English translation} (\newcite{dhar-etal-2018-enabling}, \newcite{srivastava-singh-2020-phinc})}}  \\ 
\textcolor{teal}{\textbf{Hinglish}}: Hi there! 	\colorbox{cyan!10}{Chat ke liye} ready \colorbox{cyan!10}{ho?}  & $\rightarrow$ & \textcolor{purple}{\textbf{English}}: Hi there! Ready to chat? \\
\textcolor{teal}{\textbf{Hinglish}}: \colorbox{cyan!10}{isme} kids \colorbox{cyan!10}{keliye ache} message \colorbox{cyan!10}{hein, jo} respectful \colorbox{cyan!10}{hein sabhi keliye} 
& $\rightarrow$ & \textcolor{purple}{\textbf{English}}: It does show a really good message for kids, to be respectful of everybody \\ \hline 
\multicolumn{3}{c}{\colorbox{red!10}{\textbf{English to Hinglish translation (our task)} } }  \\
\textcolor{purple}{\textbf{English}}: Maybe it’s to teach kids to challenge themselves & $\rightarrow$ & \textcolor{teal}{\textbf{Hinglish}}: maybe kida \colorbox{cyan!10}{ko} teach \colorbox{cyan!10}{karna unka} challenge \colorbox{cyan!10}{ho saktha hein} \\
\textcolor{purple}{\textbf{English}}: It’s seem to do OK on rotten tomatoes I got a 79\% & $\rightarrow$ & \textcolor{teal}{\textbf{Hinglish}}: ROTTEN TOMATOES \colorbox{cyan!10}{KA} SCORE 79\% \colorbox{cyan!10}{DIYA HEIN JO} OK \colorbox{cyan!10}{HEIN KYA} \\ \hline 
\end{tabular}
\caption{Sample translation pairs for Hinglish to English and English to Hinglish machine translation task. Words highlighted in cyan color are in Hindi language in roman script, while non-highlighted words are in English language.}
\label{fig:egs}
\end{table*}

Code-mixing is a phenomenon of mixing two or more languages in speech and text~\cite{gumperz_1982}. Code-mixing is prevalent in multilingual societies, where the speakers typically have similar fluency in two or more languages~\cite{sitaram_survey19}. For example, Hindi, Tamil and Telugu speakers from India frequently code-mix with English. Code-mixing can happen between dialects, for example, Modern Standard Arabic is frequently code-mixed with Arabic dialects~\cite{mageed2020micro}. Building NLP systems that can handle code-mixing is challenging as the space of valid grammatical and lexical configurations can be large due to presence of syntactic structures from more than one linguistic system~\cite{pratapa-etal-2018-language}.

In this work, we focus on building a machine translation (MT) system that converts a monolingual sequence of words into a code-mixed sequence. More specifically, we focus on translating from English to Hindi code-mixed with English (i.e., Hinglish). In the literature, work has been done on translating from Hinglish into English~\cite{dhar-etal-2018-enabling,srivastava-singh-2020-phinc}. To illustrate both directions, we provide Figure~\ref{fig:egs}. The Figure presents sample translation pairs for Hinglish to English as well as English to Hinglish. The challenges for solving this task include: (i) lack of Hindi data in roman script (words highlighted in cyan color), (ii) non-standard spellings (e.g., `isme' vs `is me'), (iii) token-level ambiguity across the two languages (e.g., Hindi `main' vs. English `main'), (iv) non-standard casing (e.g., ROTTEN TOMATOES), (v) informal writing style, and (vi) paucity of English-Hinglish parallel data. Compared with Hinglish to English translation, the English to Hinglish translation direction is a less studied research problem.

English-Hinglish translation can have several practical applications. For example, it can be used to create engaging conversational agents that mimic the code-mixing norm of a human user who uses code-mixing. Another use of resulting Hinglish data would be to create training data for some downstream applications such as token-level language identification. 

Our proposed machine translation system exploits a multilingual text-to-text Transformer model along with synthetically generated code-mixed data. More specifically, our system utilizes the state-of-the-art pre-trained multilingual generative model, mT5 (a multilingual variant of ``Text-to-Text Transfer Transformer" model~\cite{t5}) as a backbone. The mT5 model is pretrained on large amounts of monolingual text from $107$ languages, making it a good starting point for multilingual applications such as question answering and MT. It is not clear, however, how mT5's representation fares in a code-mixed context such as ours. {\sl This is the question we explore, empirically, in this paper, on our data.} We also introduce a simple approach for generating code-mixed data and show that by explicitly finetuning the model on this code-mixed data we are able to acquire sizeable improvements. For this finetuning, we adopt a \textit{curriculum learning} method, wherein the model is finetuned on the synthetically generated code-mixed data and then finetuned on the gold code-mixed data. 

To synthetically generate code-mixed data, we propose a novel lexical substitution method that exploits bilingual word embeddings trained on shuffled context obtained from English-Hindi bitext. The method works by replacing select $n$-grams in English sentences with their Hindi counterparts obtained from the bilingual word embedding space. For meaningful comparisons, we also 
experiment with five different methods to create code-mixed training data: (i) \textit{romanization of monolingual Hindi} from English-Hindi parallel data, (ii) \textit{paraphrasing of monolingual English} from English-Hinglish parallel data, (iii) \textit{backtranslation} of output from the mT5 model trained on English-Hinglish parallel data, (iv) \textit{adapting social media data} containing parallel English-Hinglish sentences by removing emoticons, hashtags, mentions, URLs~\cite{srivastava-singh-2020-phinc}, and (v) code-mixed data generated based on \textit{equivalence constraint theory}~\cite{pratapa-etal-2018-language}. We study the impact of different settings (e.g., size of training data, number of paraphrases per input) applicable for most methods on the translation performance. We observe that the mT5 model finetuned on the code-mixed data generated by our proposed method based on bilingual word embeddings followed by finetuning on gold data achieves a BLEU score of 12.67 and places us first in the overall ranking for the shared task. Overall, our major contributions are as follows:
\begin{enumerate}
    \item We propose a simple, yet effective and dependency-free, method to generate English-Hinglish parallel data by leveraging bilingual word embeddings trained on shuffled context obtained via English-Hindi bitext.
    \item We study the effect of several data augmentation methods (based on romanization, paraphrasing, backtranslation, etc.) on the translation performance.
    \item Exploiting our code-mixing generation method in the context of curriculum learning, we obtain state-of-the-art performance on the English-Hinglish shared task data with a BLEU score of $12.67$.
\end{enumerate}

\section{Related Work}
\label{sec:relwork}
Our work involves code-mixed data generation, machine translation involving code-mixed language, and multilingual pretrained generative models.

\subsection{Code-Mixed Data Generation}
Due to the paucity of code-mixed data, researchers have developed various methods to automatically generate code-mixed data. An ideal method for code-mixed data generation should aim to generate \textit{syntactically valid} (i.e., fluent), semantically correct words (i.e., adequate), \textit{diverse} code-mixed data of \textit{varying lengths}. To create grammatically valid code-mixed sentences, \newcite{pratapa-etal-2018-language} leverages a linguistically motivated technique based on equivalence constraint theory~\cite{POPLACK+1980+581+618}. They observe that the default distribution of synthetic code-mixed sentences created by their method can be quite different from the distribution of real code-mixed sentences in terms of code-mixing measures. This distribution gap can be largely bridged by post-processing the generated code-mixed sentences by binning them into switch point fraction bins and appropriately sampling from these bins. However, the method depends on availability of a word alignment model, which can be erroneous for distant languages (e.g., Hindi and Chinese)~\cite{gupta-etal-2020-semi}. \newcite{winata-etal-2019-code} show that a Seq2Seq model with a copy mechanism can be trained to consume parallel monolingual data (concatenated) as input and produce code-mixed data as output, that is distributionally similar to real code-mixed data. Their method needs an external NMT system to obtain monolingual fragment from code-switched text and is expensive to scale to more language pairs. \newcite{garg-etal-2018-code} introduces a novel RNN unit for an RNN based language model that includes separate components to focus on each language in code-switched text. They utilize training data generated from SeqGAN along with syntactic features (e.g., Part-of-Speech tags, Brown word clusters, language ID feature) to train their RNN based language model. Their method involves added cost to train SeqGAN model and expensive to scale to more language pairs.

\newcite{samanta_ijcai19} propose a two-level hierarchical variational autoencoder that models syntactic signals in the lower layer and language switching signals in the upper layer. Their model can leverage modest real code-switched text and large monolingual text to generate large amounts of code-switched text along with its language at token level. The code-mixed data generated by their model seems syntactically valid, yet distributionally different from real code-mixed data and their model is harder to scale for large training data. \newcite{gupta-etal-2020-semi} proposes a two-phase approach: (i)  creation of  synthetic code-mixed sentences from monolingual bitexts (English being one of the languages) by replacing aligned named entities and noun phrases from English; and (ii) training a Seq2Seq model to take English sentence as input and produce the code-mixed sentences created in the first phase. Their approach depends on the availability of a word alignment tool, a part-of-speech tagger, and knowledge of what constituents to replace in order to create a code-mixed sentence. By contrast, our proposed method based on bilingual word embeddings to generate code-mixed data does not require external software such as a word alignment tool, part-of-speech tagger, or constituency parser. 
\newcite{rizvi-etal-2021-gcm} develops the toolkit for code-mixed data generation for a given language pair using two linguistic theories: equivalence constraint (code-mixing following the grammatical structure of both the languages) and matrix language theory~\cite{mcclure_1995} (code-mixing by fixing a language that lends grammatical structure while other language lends its vocabulary). For comparison, we use this tool to implement the code-mixed data generation method based on equivalence constraint theory.
 
\subsection{Code-Mixed MT}
Building MT systems involving code-mixed language is a less researched area. Existing MT systems trained on monolingual data fail to translate code-mixed language such as from Hinglish to English~\cite{dhar-etal-2018-enabling,srivastava-singh-2020-phinc}. Given that neural MT systems require large training data, \newcite{dhar-etal-2018-enabling} collects a parallel corpus of $6,096$ Hinglish-English bitexts. They propose a machine translation pipeline where they first identify the languages involved in the code-mixed sentence, determine the matrix language, translate the longest word sequence belonging to the embedded language to the matrix language, and then translate the resulting sentence into the target language. The last two steps are performed by monolingual translation systems trained to translate embedded language to matrix language and matrix language to target language respectively. Their proposed pipeline improves the performance of Google and Bing translation systems. \newcite{srivastava-singh-2020-phinc} collect a large parallel corpus (called PHINC) of $13,738$ Hinglish-English bitexts that they claim is topically diverse and has better annotation quality than the corpus collected by \newcite{dhar-etal-2018-enabling}. They propose a translation pipeline where they perform token level language identification and translate select phrases involving mostly Hindi to English using a monolingual translation system, while keeping the rest of phrases intact. This proposed pipeline outperforms Google and Bing systems on the PHINC dataset. For our work, we make use of the PHINC dataset by adapting the text by removing mentions, hashtags, emojis, emoticons as well as non-meaning bearing constituents such as URLs.

\subsection{Multilingual Pretrained Models}
Neural models pretrained on monolingual data using a self-supervised objective such as BERT~\cite{devlin_naacl19}, BART~\cite{bart}, and T5~\cite{t5} have become integral to NLP systems as they serve as a good starting point for building SOTA models for diverse monolingual tasks. Recently, there is increasing attention to pretraining neural models on multilingual data, resulting in models such as mBERT~\cite{devlin_naacl19}, XLM~\cite{conneau2019unsupervised}, mBART~\cite{mbart} and mT5~\cite{mt5}. Especially, generative multilingual models such as mBART~\cite{mbart} and mT5~\cite{mt5} can be utilized directly without additional neural network components to solve summarization, MT, and other natural language generation tasks. These generative models are trained using a self-supervised pretraining objective based on span-corruption objective (mBART and mT5) and sentence shuffling objective (mBART). Training data for these models are prepared by concatenating monolingual texts from multiple languages (e.g., 25 for mBART, 107 for mT5). It is not clear how much code-mixed data these models have seen during pretraining, making it an important question to investigate how they fare in processing text in varieties such as Hinglish. In this work, we target this question by exploring the challenges of applying one of these models (mT5) for the English to Hinglish translation task.

\section{Shared Task}
\label{sec:shtask}
The goal of the shared task is to encourage MT involving code-mixing. We focus on translating English to Hinglish. A sentence in Hinglish may contain English tokens and roman Hindi tokens, as shown in Figure~\ref{fig:egs}. 
The organizers provide $8,060$, $942$ and $960$ examples for training, validation, and test  respectively. 

\section{Our Approach}
\label{sec:ourappr}

Our approach to the English-Hinglish MT task is simple. We first identify the best text-to-text Transformer model on the validation set and follow a curriculum learning procedure to finetune the model for the downstream task. The curriculum learning procedure works such that we first finetune the model using synthetic code-mixed data from our generation method, then further finetune on the gold code-mixed data.  This training recipe has been explored previously by \newcite{choudhury-etal-2017-curriculum} and \newcite{pratapa-etal-2018-language} to build code-mixed language models. Curriculum learning itself has been explored previously for different NLP tasks such as parsing~\cite{spitkovsky-etal-2010-baby} and language modeling~\cite{graves2017automated}. We now present our proposed method to \textit{generate} synthetic code-mixed text for a given language pair. 


For our method, we assume having access to large amounts of bitext from a given pair of languages ($LG_1$ and $LG_2$) for which we need to generate code-mixed data. Let $B_i=\{x_i,y_i\}$ denote the bitext data, where $x_i$ and $y_i$ correspond to sentences in $LG_1$ and $LG_2$, respectively. Let $\texttt{\mbox{ngrams}}(n,x_i,y_i)$ denote the set of unique $n$-grams in $x_i$ and $y_i$. Let $\texttt{\mbox{cumulative-ngrams}}(n,x_i,y_i) = \cup_{j=1}^{j=n} \texttt{\mbox{ngrams}}(j,x_i,y_i)$ denote the cumulative set of unique $n$-grams in the set of pairs $x_i$ and $y_i$. We shuffle the $n$-grams in the cumulative set and create a ``shuffled'' code-mixed sentence by concatenating the shuffled set with $n$-grams separated by a space. For example, let $LG_1$ denote English and $LG_2$ denote Hindi (assuming Roman script for illustration). A sample bitext instance $B_i$ can be ``I've never seen it'' ($x_i$) and ``maine ye kabhi nah dekhi'' ($y_i$). Set of unique $1$-grams will be $\{$``I've'', ``never'', ``seen'', ``it'', ``maine'', ``ye'', ``kabhi'', ``nah'', ``dekhi''$\}$ ($\texttt{\mbox{ngrams}}(1,x_i,y_i)$, assuming a whitespace tokenizer for simplicity). Then,  $\texttt{\mbox{cumulative-ngrams}}(2,x_i,y_i)$ correspond to $\{$``I've'', ``never'', ``seen'', ``it'', ``maine'', ``ye'', ``kabhi'', ``nah'', ``dekhi'', ``I've never'', ``never seen'', ``seen it'', ``maine ye'', ``ye kabhi'', ``kabhi nah'', ``nah dekhi''$\}$. A shuffled code-mixed sentence can be, ``I've ye\_kabhi never seen\_it seen never\_seen it kabhi\_nah I've\_never  maine\_ye ye kabhi nah dekhi maine nah\_dekhi''. We create  one shuffled code-mixed sentence per bitext instance, thereby creating a shuffled code-mixed corpus. We train a word2vec model on this shuffled code-mixed corpus to learn embeddings for $n$-grams in both languages. The resulting word embeddings seem cross-lingually aligned (based on manual inspection), thereby allowing us to do $n$-gram translation from one language to another language. For example, nearest English neighbor of a Hindi $1$-gram ``nah'' can be ``never''. 

Once the word embeddings are learned, we can create a code-mixed sentence for the given languages: $LG_1$ and $LG_2$. We first find the $n$-grams in $x_i \in LG_1$ and then sort all the $n$-grams by cosine similarity of the $n$-gram with its most similar $n$-gram in $LG_2$. Let $\texttt{\mbox{num-substitutions}}$ denote the number of substitutions performed to convert $x_i$ to a code-mixed sentence. We pick one $n$-gram at a time from the sorted list and replace all occurrences of that $n$-gram with its top $n$-gram belonging to language $LG_2$ based on word embeddings. We continue this substitution process until we exhaust the $\texttt{\mbox{num-substitutions}}$. 

For our machine translation task, we assume $LG_1$ and $LG_2$ to be English and Hindi (native) respectively.~\footnote{We can assume $LG_1$ and $LG_2$ to be Hindi and English respectively, but we leave this exploration for future.} We feed the OPUS corpus~\footnote{\url{https://opus.nlpl.eu/}} containing 17.2M English-Hindi bitexts (Hindi in native script) as input to the algorithm that outputs English-Hinglish code-mixed parallel data.

\section{Experiments}
\label{sec:experiments}

\begin{table}[]
\footnotesize
\centering
\begin{tabular}{p{1.1in}HHp{0.4in}p{0.4in}}  \hline
cs method (hyper.) & S1 epoch &  S2 epoch & Valid & Test \\ \hline
\textit{Romanization}  \\ 
IIT-B (100K) & 3 & 50 & 14.27 & 12.95 \\
IIT-B (250K) & 5 & 47 & \textbf{14.74} & \textbf{12.75} \\
IIT-B (500K) & 3 & 46 & 14.12 & 12.46 \\ 
OPUS (100K) & 3 & 43 & 14.67 & 12.62 \\
OPUS (250K) & 3 & 50 & 14.57 & 12.71 \\ \hline
\textit{Paraphrasing}  \\ 
Para (1) & 5 & 43 & 14.39 & 12.72 \\
Para (2) & 5 & 43 & 14.4 & 12.62 \\
Para (3) & 5 & 44 & \textbf{15.07} & \textbf{12.63} \\ \hline
\textit{Backtranslation} \\
Forward model & 3 & 37 & \textbf{14.07} & \textbf{12.16} \\
Backward model & 3 & 36 & \textbf{14.51} & \textbf{13.03} \\ \hline
\textit{Social media} \\
PHINC & 5 & 29 & \textbf{14.71} & \textbf{12.68} \\ \hline
\textit{CMDR (ours)}
 \\ 
CMDR-unigram & 3 & 48 & \textbf{14.6} & \textbf{12.69} \\
CMDR-bigram & 5 & 42 & 14.58 & 12.4 \\ \hline
\end{tabular}
\caption{Performance in BLEU of mT5 model finetuned using curriculum learning --- finetuning on one of the different code-mixed data generation method followed by finetuning on gold data. \textbf{CMDR:} Code-Mixing from Distributed Representations refers to our proposed method. Note that we did not study the method based on equivalence constraint theory in this experiment. For CMDR, we perform $n$-gram translation of Hindi from native to roman script.} 
\label{tab:cus_splits}
\end{table}

In this section, we first discuss how we choose a text-to-text Transformer model from available models and then introduce our five baseline methods.

\subsection{Choosing a Text-to-Text Transformer}
\label{sec:choosing_plm}
Multilingual encoder-decoder models such as mT5~\cite{mt5}\footnote{\url{https://github.com/google-research/multilingual-t5}} and 
mBART~\cite{mbart}\footnote{\url{https://github.com/pytorch/fairseq/tree/master/examples/mbart}} are suited to the MT task, and already cover both English and Hindi. It is not clear, however, how these models will perform on a task involving code-mixing at the target side such as ours (where we need to output Hinglish). For this reason, we first explore the potential of these two models on the code-mixed translation task to select the best model among these two. Once we identify the best model, we use it as the basis for further experiments as we will explain in Section~\ref{res:cus_splits}. For both mT5 and mBART, we use the implementation provided by the HuggingFace library~\cite{wolf-etal-2020-transformers} with the default settings for all hyperparameters except the maximum number of training epochs, which we choose based on the validation set. 

\begin{figure}[ht]
\centering
\includegraphics[width=3.1in, height=1.3in]{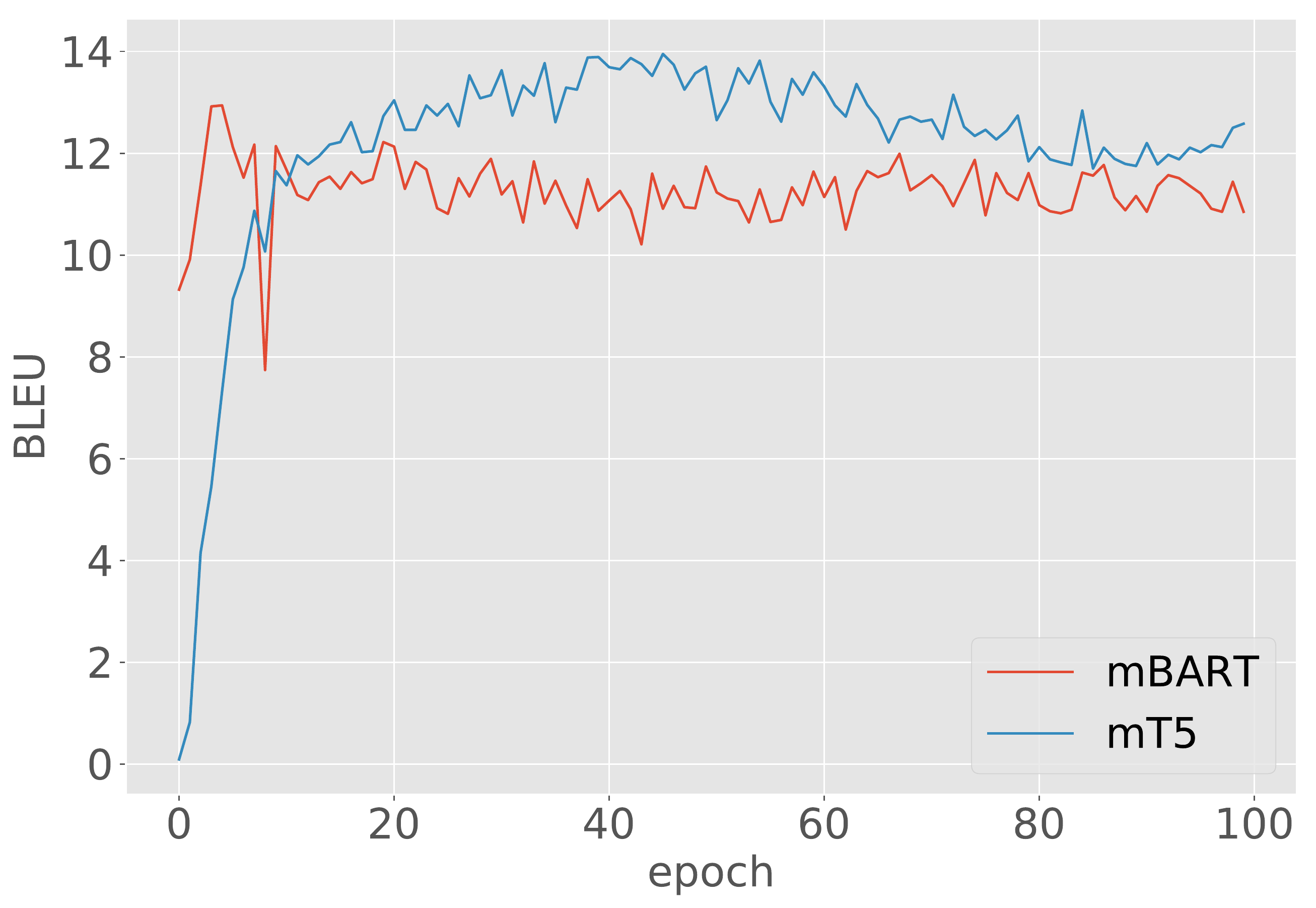}
\caption{Validation BLEU of mBART and mT5 model on 541 randomly picked examples from the official training set after deduplication, while the rest of the 7000 examples are used for training.}
\label{fig:det_acc}
\end{figure}

\textbf{Data Splits.} For this set of experiments, we 
use ``custom'' splits using the official shared task training data after deduplication\footnote{Deduplication is done based on exact overlap of source and target text.} and shuffling, as follows: $7,000$ examples for training set and $541$ examples for validation set. For testing test, we use the official validation data (n=$942$ examples).
We finetune both mT5 and mBART on the custom split, and show results in Figure~\ref{fig:det_acc}. We observe that mBART converges quickly within $5$ epochs, while mT5 model takes $\sim 46$ epochs for convergence. Importantly, the best validation performance of 13.95 BLEU is obtained by the mT5 model, which helped us choose mT5 as the backbone model to build our final MT system. For subsequent experiments, we choose $50$ as the maximum number of epochs to finetune the mT5 model. We now introduce our baseline code-mixing data generation methods.

\subsection{Baseline Code-Mixing Generation Methods}
\label{sec:baseline_cs_methods}
We experiment with five different baseline methods to generate English-Hinglish bitexts that can be used in the first stage of finetuning. We now describe each of these methods.

\subsubsection{Monolingual Target Romanization}
\label{sec:romanization}
In this method, we focus on creating monolingual bitexts by taking the Hindi sentence from  parallel English-Hindi data and changing the script of Hindi from native script (Devanagari) to Roman script while keeping the English sentence intact. Although the resulting Hindi sentence is monolingual, the generated bitexts can help mT5 model to learn the semantics of Romanized Hindi language (mT5 model might be pretrained on native Hindi), along with the relationships between English and romanized Hindi language. To this end, we exploit two large parallel data sources for English-Hindi pairs (Hindi in native script) ---  IIT Bombay Parallel corpus~\cite{kunchukuttan-etal-2018-iit} (1.49M bitexts) and OPUS corpus (17.2M bitexts). We utilize the Aksharamukha tool to convert native Hindi to romanized Hindi.\footnote{\url{https://aksharamukha.appspot.com}}

\subsubsection{Monolingual Source Paraphrasing}
\label{sec:paraphrasing}
Here, we paraphrase each English source sentence in the gold data to create a new training example, while keeping the target Hinglish sentence intact. Since good paraphrases can typically retain the meaning of the original sentence although the form can be different, we hypothesize the resulting bitext can improve the robustness of our translation system. To generate paraphrases, we use the T5\textsubscript{BASE}~\cite{t5} model finetuned on paraphrases from diverse English sources. For our experiments, we use \textit{n} paraphrases of each source sentence, with \textit{n} chosen from the set \{{1,2,3}\}. Details about our paraphrasing model are in Appendix~\ref{sec:msp}. 


\subsubsection{Backtranslation}
\label{sec:btrans}
We also use the traditional backtranslation pipeline to generate more data for our task. Specifically, we create two models: \textit{forward model} that is obtained by finetuning the mT5 model on English as source and Hinglish as target, \textit{backward model} that is obtained by finetuning mT5 on Hinglish as source and English as target, on the gold training data in both cases. For each gold bitext, the process involves two steps: \textit{forward model inference}, where the gold English sentence is fed to the forward model that generates the intermediate Hinglish sentence; \textit{backward model inference}, where the intermediate Hinglish sentence is fed to the backward model that generates the final English sentence. The new bitext is obtained by pairing up the final English sentence with the gold Hinglish sentence (which is parallel to the English fed to the forward model as source). This method can be treated as an alternative method to creating paraphrases of an English sentence.

\subsubsection{Social Media Adaptation}
\label{sec:clean_smedia}
We adapt a publicly available English-Hinglish social media dataset, PHINC~\cite{srivastava-singh-2020-phinc}, to our task. PHINC consists of $13,738$ manually annotated English-Hinglish code-mixed sentences, mainly sourced from social media platforms such as Twitter and Facebook. It covers a wide range of topics (such as sports and politics) and has high quality text (e.g., it handles spelling variations and filters abusive and ambiguous sentences). We perform post-processing on PHINC by removing tokens particular to the social media context such as hashtags, mentions, emojis, emoticons and URLs. We use the resulting, adapted, dataset to finetune mT5 for the first stage (as explained in Section~\ref{sec:ourappr}).

\subsubsection{Equivalence Constraint Theory}
\label{sec:ect}
This method generates code-mixed data based on \textit{equivalence constraint theory} (EC), as originally proposed by \newcite{pratapa-etal-2018-language}. The method works by producing parse trees for English-Hindi sentence pair and replaces common nodes between the two trees based on the EC theory. We use the implementation provided by the GCM tool~\cite{rizvi-etal-2021-gcm}. We feed the English-Hindi bitexts (Hindi in native script) from the OPUS corpus to generate English-Hinglish (Hindi in native script) parallel data. We now describe our results with mT5 on our custom splits.

\subsection{Performance With mT5}
\label{res:cus_splits}
As briefly introduced earlier, we finetune the mT5 model using curriculum learning where we have two stages. In stage one, we finetune one of the code-mixed data generation methods. We follow that by stage two where we finetune on the gold code-mixed data (official shared task training data). Also, for stage one, to cap GPU hours with the large synthetic code-mixed data, we experiment with a maximum of 5 epochs. 
For the stage two, where we have smaller amount of gold data, we experiment with 50 as the maximum number of epochs choosing the best epoch on the validation set. 

Table~\ref{tab:cus_splits} displays the validation and the test performance of mT5 finetuned using curriculum learning.~\footnote{The best epoch for each stage in the pipeline is displayed in Appendix~\ref{sec:appd_mt5_perf}.} For romanization of monolingual target method, as the Table shows, more data does not strictly improve validation (nor test) performance. That is, there seems to be a `sweet spot' after which quality deteriorates with noise. The behavior is similar for the models exploiting paraphrases of the source monolingual English data: Adding more paraphrases for a single gold instance can lead to overfitting of the model, as noticed by consistent degradation in test performance. For backtranslation, we experiment with two variants: \textit{forward model} where predictions (Hinglish) from the forward model is paired up with English sentence from the gold, \textit{backward model} which corresponds to the traditional backtranslation bitext. Performance of the backward model is consistently better on both the validation and the test set. For the social media adaptation method, mT5 achieves validation performance that is better than any of the methods based on romanization or backtranslation. For our proposed method based on code-mixing from bilingual distributed representations (CMDR), we experiment with different values of \texttt{num-substitutions} and change the script of replaced Hindi words from native to roman script using the Aksharamukha tool. Manual inspection of the data reveals that script conversion at word level is noisy due to lack of sentential context. This might lead to decline in the performance as our method makes more substitutions. Nevertheless, our proposed method, simple as it is, leads to results competitive with any of the other methods.

\subsection{Qualitative Analysis}
\label{sec:qa}
We manually inspect translations from our proposed system that uses native script and 3 as \texttt{num-substitutions} on 25 randomly picked examples from the official test set. 64\% of the translations are correct, while 24\% and 12\% of the translations have grammatical error (e.g., incorrect gender) and semantic errors (e.g., factual inconsistency) respectively. 12\% of the translations exactly match with the source. Few of these translations are shown in Table~\ref{tab:qa}. The first translation has grammatical gender error, as it contains male posessive noun, `ka' (instead of female possessive noun, `ki'). The second translation has semantic error, where the number of times that the movie has been watched is incorrectly translated as one time (`ek') when the source mentions it as two (`do') times. The third example is long (43 words), which our system translates without errors.


\begin{table}[]
\footnotesize
\centering
\begin{tabular}{p{3in}} \hline
\textbf{English (Gold)}: And they grow apart. She is the protector of the Moors forest. \\ 
\textbf{Hinglish (Prediction)}: Aur wo apart grow karte hai. Wo Moors forest \colorbox{red!20}{ka} (\colorbox{green!20}{ki}) protector hai. \\  \hline
\textbf{English (Gold)}: I watched it at least twice.. it was that good.  I love female superheros \\ 
\textbf{Hinglish (Prediction)}: Maine ise kam se kam \colorbox{red!20}{ek} (\colorbox{green!20}{do}) baar dekha hai. Ye itni achi thi. Mujhe female superheros pasand hai. \\  \hline
\textbf{English (Gold)}: I loved the story \& how true they made it in how  teen girls act but I honestly don't know why I didn't rate it highly as all the right ingredients were there. I cannot believe it was 2004 it was released though, 14 years ago! \\
\textbf{Hinglish (Prediction)}: mujhe story bahut pasand aaya aur teen girls ka act kaise hota lekin main honestly nahi janta kyon ki main ise highly rate nahi kar raha tha kyunki sahi ingredients wahan they. mujhe yakin nahi hota ki 2004 mein release huyi thi, 14 saal pehle! \\ \hline
\end{tabular}
\caption{Translations of our proposed system that uses native script and 3 as \texttt{num-substitutions}. Errors in translations are highligthed in red color, with their the right translation in paranthesis and highlighted in green color.}
\label{tab:qa}
\end{table}

\section{Official Results}
\label{sec:res_official}

\begin{table}[]
\footnotesize
\centering
\begin{tabular}{p{2.4in}p{0.3in}}  \hline
cs method & BLEU \\ \hline
baseline (mBART model) & 11.00 \\ \hline
\textit{LinCE leaderboard} (only best results)   \\ 
LTRC Team & 12.22 \\ 
IITP-MT Team & 10.09 \\
CMMTOne Team & 2.58 \\ \hline
\textit{Romanization}  \\ 
OPUS & 12.38 \\ \hline
\textit{Paraphrasing}  \\
Para & 12.1 \\ \hline
\textit{Backtranslation} \\
Backward model & 11.47 \\ \hline
\textit{Social media} \\
PHINC & 11.9 \\ \hline
\textit{Equivalence constraint theory } \\ 
ECT (100K) & 12.45 \\ \hline
\textit{CMDR (ours)} \\
CMDR-unigram (roman) & 12.25 \\
CMDR-bigram (native) & 12.63 \\
CMDR-bigram (roman) & 12.08 \\
CMDR-trigram (native) &\textbf{12.67} \\
CMDR-trigram (roman) & 12.05 \\ \hline
\textit{Method Combinations} \\
CMDR-unigram (roman) + PHINC  & 11.58 \\
ECT (100K) + CMDR-trigram (native) & 12.27 \\ \hline
\end{tabular}
\caption{Official test performance of mT5 model finetuned using curriculum learning --- finetuning on one of the different code-mixed data generation method (max. epochs is  5) followed by finetuning on concatenation of gold training data and gold validation data (leaving out 200 examples for validation) (max. epochs is 50)}
\label{tab:off_splits}
\end{table}

In this section, we describe the official test performance obtained by our models. First, we experiment with mT5 model finetuned using promising code-mixing methods identified in our previous experiments (see Section~\ref{res:cus_splits}). The best performing baseline method is based on equivalence constraint theory for $100K$ examples and yields a BLEU score of 12.45. For the proposed CMDR method, we experiment not only with the value for \texttt{num-substitutions}, but also the script type. Surprisingly, the best combination for our proposed method is based on maximum substitutions of $3$,  sticking to the original native script, and yields the highest BLEU score of $12.67$. The variants of our proposed method that romanizes the replacement n-gram consistently perform poorly, which confirms our observation that n-gram level romanization is deprived of sentential context and is prone to errors.

\note[mam]{Very good!}

\section{Discussion}
\label{sec:discussion}



The lessons learned in this shared task can be summarized as follows. 
    
\noindent\textbf{Similar Text-to-Text Models.} \textit{Off-the-shelf mT5 and mBART models perform similarly, with mT5 being slightly better in our experiments (for English-Hinglish MT)}. A down side of mT5 is that it takes many more epochs than mBART to converge. In the future, it will be interesting to explore recent extensions of mBART\footnote{These are \texttt{mbart-large-50-many-to-many-mmt}, \texttt{mbart-large-50-one-to-many-mmt}, and \texttt{mbart-large-50-many-to-one-mmt}.}, which are already finetuned for multilingual translation. These extensions involve training on English-Hindi (native) bitexts, and so can act as an interesting zero-shot translation baseline without further finetuning. They may also serve as a better baseline when finetuned using the curriculum learning approach adopted in our work.

\noindent\textbf{Code-Mixing from Distributed Representations is Useful.} \textit{Our proposed code-mixed data generation method based on bilingual word embeddings can be exploited by mT5 model to achieve the state-of-the-art translation performance,} especially when the number of substitutions is high and the script remains in native form. It will be interesting to see the sweet spot for the number of substitutions, as  too low value can result in very less code-mixing while too high value can result in more code-mixing along with more noise (possibly grammatically incorrect and unnatural to bilingual speaker).

\noindent\textbf{Combinations of Code-Mixed Data not Ideal.} \textit{Combining code-mixed generations from two methods likely introduces more noise and does not improve the performance of the mT5 model compared to performance obtained using generations from individual method}, as seen in the `Misc.' section of Table~\ref{tab:off_splits}. It might be interesting to explore more than two stages of curriculum learning, where the mT5 model is successively finetuned on code-mixed data generated using different methods.

\section{Conclusion}
\label{sec:conclusion}
\note[mam]{I stopped here. I also didn't review the abstract yet--left it to the end.}
We proposed an MT pipeline for translating between English and Hinglish. We test the utility of existing pretrained language models on the task and propose a simple, dependency-free, method for generating synthetic code-mixed text from bilingual distributed representations of words and phrases. Comparing our proposed method to five baseline methods, we show that our method achieves competitively. The method results in best translation performance on the shared task blind test data, placing us first in the official competition. In the future, we plan to (i) scale up the size of code-mixed data, (ii) experiment with different domains of English-Hindi bitexts such as Twitter, (iii) experiment with recent extensions of mBART, and (iv) assess the generalizability of our proposed code-mixing method to other NLP tasks such as question answering and dialogue modeling.
\\


\section*{Acknowledgements}
\label{sec:acknow}
We gratefully acknowledges support from the Natural Sciences and Engineering Research Council of Canada, the Social Sciences and Humanities Research Council of Canada, Canadian Foundation for Innovation, Compute Canada (\url{www.computecanada.ca}), and UBC ARC-Sockeye (\url{https://doi.org/10.14288/SOCKEYE}).

\bibliography{anthology,custom}

\begin{thebibliography}{28}
\expandafter\ifx\csname natexlab\endcsname\relax\def\natexlab#1{#1}\fi

\bibitem[{Abdul-Mageed et~al.(2020)Abdul-Mageed, Zhang, Elmadany, and
  Ungar}]{mageed2020micro}
Muhammad Abdul-Mageed, Chiyu Zhang, AbdelRahim Elmadany, and Lyle Ungar. 2020.
\newblock Micro-dialect identification in diaglossic and code-switched
  environments.
\newblock In \emph{Proceedings of the 2020 Conference on Empirical Methods in
  Natural Language Processing (EMNLP)}, pages 5855--5876.

\bibitem[{Choudhury et~al.(2017)Choudhury, Bali, Sitaram, and
  Baheti}]{choudhury-etal-2017-curriculum}
Monojit Choudhury, Kalika Bali, Sunayana Sitaram, and Ashutosh Baheti. 2017.
\newblock \href {https://www.aclweb.org/anthology/W17-7509} {Curriculum design
  for code-switching: Experiments with language identification and language
  modeling with deep neural networks}.
\newblock In \emph{Proceedings of the 14th International Conference on Natural
  Language Processing ({ICON}-2017)}, pages 65--74, Kolkata, India. NLP
  Association of India.

\bibitem[{Conneau et~al.(2019)Conneau, Khandelwal, Goyal, Chaudhary, Wenzek,
  Guzm{\'a}n, Grave, Ott, Zettlemoyer, and Stoyanov}]{conneau2019unsupervised}
Alexis Conneau, Kartikay Khandelwal, Naman Goyal, Vishrav Chaudhary, Guillaume
  Wenzek, Francisco Guzm{\'a}n, Edouard Grave, Myle Ott, Luke Zettlemoyer, and
  Veselin Stoyanov. 2019.
\newblock Unsupervised cross-lingual representation learning at scale.
\newblock \emph{arXiv preprint arXiv:1911.02116}.

\bibitem[{Creutz(2018)}]{creutz2018open}
Mathias Creutz. 2018.
\newblock Open subtitles paraphrase corpus for six languages.
\newblock \emph{arXiv preprint arXiv:1809.06142}.

\bibitem[{Devlin et~al.(2019)Devlin, Chang, Lee, and
  Toutanova}]{devlin_naacl19}
Jacob Devlin, Ming-Wei Chang, Kenton Lee, and Kristina Toutanova. 2019.
\newblock {{{BERT}: Pre-training of Deep Bidirectional Transformers for
  Language Understanding}}.
\newblock In \emph{Proceedings of the 2019 Conference of the North {A}merican
  Chapter of the Association for Computational Linguistics: Human Language
  Technologies, Volume 1 (Long and Short Papers)}, pages 4171--4186.

\bibitem[{Dhar et~al.(2018)Dhar, Kumar, and
  Shrivastava}]{dhar-etal-2018-enabling}
Mrinal Dhar, Vaibhav Kumar, and Manish Shrivastava. 2018.
\newblock \href {https://www.aclweb.org/anthology/W18-3817} {Enabling
  code-mixed translation: Parallel corpus creation and {MT} augmentation
  approach}.
\newblock In \emph{Proceedings of the First Workshop on Linguistic Resources
  for Natural Language Processing}, pages 131--140, Santa Fe, New Mexico, USA.
  Association for Computational Linguistics.

\bibitem[{Garg et~al.(2018)Garg, Parekh, and Jyothi}]{garg-etal-2018-code}
Saurabh Garg, Tanmay Parekh, and Preethi Jyothi. 2018.
\newblock \href {https://doi.org/10.18653/v1/D18-1346} {Code-switched language
  models using dual {RNN}s and same-source pretraining}.
\newblock In \emph{Proceedings of the 2018 Conference on Empirical Methods in
  Natural Language Processing}, pages 3078--3083, Brussels, Belgium.
  Association for Computational Linguistics.

\bibitem[{Graves et~al.(2017)Graves, Bellemare, Menick, Munos, and
  Kavukcuoglu}]{graves2017automated}
Alex Graves, Marc~G. Bellemare, Jacob Menick, Remi Munos, and Koray
  Kavukcuoglu. 2017.
\newblock \href {http://arxiv.org/abs/1704.03003} {Automated curriculum
  learning for neural networks}.

\bibitem[{Gumperz(1982)}]{gumperz_1982}
John~J. Gumperz. 1982.
\newblock \href {https://doi.org/10.1017/CBO9780511611834} {\emph{Discourse
  Strategies}}.
\newblock Studies in Interactional Sociolinguistics. Cambridge University
  Press.

\bibitem[{Gupta et~al.(2020)Gupta, Ekbal, and
  Bhattacharyya}]{gupta-etal-2020-semi}
Deepak Gupta, Asif Ekbal, and Pushpak Bhattacharyya. 2020.
\newblock \href {https://doi.org/10.18653/v1/2020.findings-emnlp.206} {A
  semi-supervised approach to generate the code-mixed text using pre-trained
  encoder and transfer learning}.
\newblock In \emph{Findings of the Association for Computational Linguistics:
  EMNLP 2020}, pages 2267--2280, Online. Association for Computational
  Linguistics.

\bibitem[{Holtzman et~al.(2020)Holtzman, Buys, Du, Forbes, and
  Choi}]{holtzman_iclr20}
Ari Holtzman, Jan Buys, Li~Du, Maxwell Forbes, and Yejin Choi. 2020.
\newblock {The Curious Case of Neural Text Degeneration}.
\newblock In \emph{International Conference on Learning Representations}.

\bibitem[{Kunchukuttan et~al.(2018)Kunchukuttan, Mehta, and
  Bhattacharyya}]{kunchukuttan-etal-2018-iit}
Anoop Kunchukuttan, Pratik Mehta, and Pushpak Bhattacharyya. 2018.
\newblock \href {https://www.aclweb.org/anthology/L18-1548} {The {IIT} {B}ombay
  {E}nglish-{H}indi parallel corpus}.
\newblock In \emph{Proceedings of the Eleventh International Conference on
  Language Resources and Evaluation ({LREC} 2018)}, Miyazaki, Japan. European
  Language Resources Association (ELRA).

\bibitem[{Lan et~al.(2017)Lan, Qiu, He, and Xu}]{lan2017continuously}
Wuwei Lan, Siyu Qiu, Hua He, and Wei Xu. 2017.
\newblock \href {https://doi.org/10.18653/v1/D17-1126} {A continuously growing
  dataset of sentential paraphrases}.
\newblock In \emph{Proceedings of the 2017 Conference on Empirical Methods in
  Natural Language Processing}, pages 1224--1234, Copenhagen, Denmark.
  Association for Computational Linguistics.

\bibitem[{Lewis et~al.(2020)Lewis, Liu, Goyal, Ghazvininejad, Mohamed, Levy,
  Stoyanov, and Zettlemoyer}]{bart}
Mike Lewis, Yinhan Liu, Naman Goyal, Marjan Ghazvininejad, Abdelrahman Mohamed,
  Omer Levy, Veselin Stoyanov, and Luke Zettlemoyer. 2020.
\newblock \href {https://doi.org/10.18653/v1/2020.acl-main.703} {{BART}:
  Denoising sequence-to-sequence pre-training for natural language generation,
  translation, and comprehension}.
\newblock In \emph{Proceedings of the 58th Annual Meeting of the Association
  for Computational Linguistics}, pages 7871--7880, Online. Association for
  Computational Linguistics.

\bibitem[{Liu et~al.(2020)Liu, Gu, Goyal, Li, Edunov, Ghazvininejad, Lewis, and
  Zettlemoyer}]{mbart}
Yinhan Liu, Jiatao Gu, Naman Goyal, Xian Li, Sergey Edunov, Marjan
  Ghazvininejad, Mike Lewis, and Luke Zettlemoyer. 2020.
\newblock \href {http://arxiv.org/abs/2001.08210} {{Multilingual Denoising
  Pre-training for Neural Machine Translation}}.

\bibitem[{McClure(1995)}]{mcclure_1995}
Erica McClure. 1995.
\newblock Duelling languages: Grammatical structure in codeswitching.
\newblock \emph{Studies in Second Language Acquisition}, 17(1):117–118.

\bibitem[{Poplack(1980)}]{POPLACK+1980+581+618}
Shana Poplack. 1980.
\newblock Sometimes i’ll start a sentence in spanish y termino en espaÑol:
  toward a typology of code-switching.
\newblock 18(7-8):581--618.

\bibitem[{Pratapa et~al.(2018)Pratapa, Bhat, Choudhury, Sitaram, Dandapat, and
  Bali}]{pratapa-etal-2018-language}
Adithya Pratapa, Gayatri Bhat, Monojit Choudhury, Sunayana Sitaram, Sandipan
  Dandapat, and Kalika Bali. 2018.
\newblock \href {https://doi.org/10.18653/v1/P18-1143} {Language modeling for
  code-mixing: The role of linguistic theory based synthetic data}.
\newblock In \emph{Proceedings of the 56th Annual Meeting of the Association
  for Computational Linguistics (Volume 1: Long Papers)}, pages 1543--1553,
  Melbourne, Australia. Association for Computational Linguistics.

\bibitem[{Raffel et~al.(2020)Raffel, Shazeer, Roberts, Lee, Narang, Matena,
  Zhou, Li, and Liu}]{t5}
Colin Raffel, Noam Shazeer, Adam Roberts, Katherine Lee, Sharan Narang, Michael
  Matena, Yanqi Zhou, Wei Li, and Peter~J. Liu. 2020.
\newblock \href {http://jmlr.org/papers/v21/20-074.html} {Exploring the limits
  of transfer learning with a unified text-to-text transformer}.
\newblock \emph{Journal of Machine Learning Research}, 21(140):1--67.

\bibitem[{Rizvi et~al.(2021)Rizvi, Srinivasan, Ganu, Choudhury, and
  Sitaram}]{rizvi-etal-2021-gcm}
Mohd Sanad~Zaki Rizvi, Anirudh Srinivasan, Tanuja Ganu, Monojit Choudhury, and
  Sunayana Sitaram. 2021.
\newblock {GCM}: A toolkit for generating synthetic code-mixed text.
\newblock In \emph{Proceedings of the 16th Conference of the European Chapter
  of the Association for Computational Linguistics: System Demonstrations},
  pages 205--211, Online. Association for Computational Linguistics.

\bibitem[{Samanta et~al.(2019)Samanta, Reddy, Jagirdar, Ganguly, and
  Chakrabarti}]{samanta_ijcai19}
Bidisha Samanta, Sharmila Reddy, Hussain Jagirdar, Niloy Ganguly, and Soumen
  Chakrabarti. 2019.
\newblock \href {http://arxiv.org/abs/1906.08972} {A deep generative model for
  code-switched text}.
\newblock \emph{CoRR}, abs/1906.08972.

\bibitem[{Sitaram et~al.(2019)Sitaram, Chandu, Rallabandi, and
  Black}]{sitaram_survey19}
Sunayana Sitaram, Khyathi~Raghavi Chandu, Sai~Krishna Rallabandi, and Alan~W.
  Black. 2019.
\newblock A survey of code-switched speech and language processing.
\newblock \emph{CoRR}, abs/1904.00784.

\bibitem[{Spitkovsky et~al.(2010)Spitkovsky, Alshawi, and
  Jurafsky}]{spitkovsky-etal-2010-baby}
Valentin~I. Spitkovsky, Hiyan Alshawi, and Daniel Jurafsky. 2010.
\newblock \href {https://www.aclweb.org/anthology/N10-1116} {From baby steps to
  leapfrog: How {``}less is more{''} in unsupervised dependency parsing}.
\newblock In \emph{Human Language Technologies: The 2010 Annual Conference of
  the North {A}merican Chapter of the Association for Computational
  Linguistics}, pages 751--759, Los Angeles, California. Association for
  Computational Linguistics.

\bibitem[{Srivastava and Singh(2020)}]{srivastava-singh-2020-phinc}
Vivek Srivastava and Mayank Singh. 2020.
\newblock \href {https://doi.org/10.18653/v1/2020.wnut-1.7} {{PHINC}: A
  parallel {H}inglish social media code-mixed corpus for machine translation}.
\newblock In \emph{Proceedings of the Sixth Workshop on Noisy User-generated
  Text (W-NUT 2020)}, pages 41--49, Online. Association for Computational
  Linguistics.

\bibitem[{Winata et~al.(2019)Winata, Madotto, Wu, and
  Fung}]{winata-etal-2019-code}
Genta~Indra Winata, Andrea Madotto, Chien-Sheng Wu, and Pascale Fung. 2019.
\newblock \href {https://doi.org/10.18653/v1/K19-1026} {Code-switched language
  models using neural based synthetic data from parallel sentences}.
\newblock In \emph{Proceedings of the 23rd Conference on Computational Natural
  Language Learning (CoNLL)}, pages 271--280, Hong Kong, China. Association for
  Computational Linguistics.

\bibitem[{Wolf et~al.(2020)Wolf, Debut, Sanh, Chaumond, Delangue, Moi, Cistac,
  Rault, Louf, Funtowicz, Davison, Shleifer, von Platen, Ma, Jernite, Plu, Xu,
  Le~Scao, Gugger, Drame, Lhoest, and Rush}]{wolf-etal-2020-transformers}
Thomas Wolf, Lysandre Debut, Victor Sanh, Julien Chaumond, Clement Delangue,
  Anthony Moi, Pierric Cistac, Tim Rault, Remi Louf, Morgan Funtowicz, Joe
  Davison, Sam Shleifer, Patrick von Platen, Clara Ma, Yacine Jernite, Julien
  Plu, Canwen Xu, Teven Le~Scao, Sylvain Gugger, Mariama Drame, Quentin Lhoest,
  and Alexander Rush. 2020.
\newblock \href {https://doi.org/10.18653/v1/2020.emnlp-demos.6} {Transformers:
  State-of-the-art natural language processing}.
\newblock In \emph{Proceedings of the 2020 Conference on Empirical Methods in
  Natural Language Processing: System Demonstrations}, pages 38--45, Online.
  Association for Computational Linguistics.

\bibitem[{Xu et~al.(2015)Xu, Callison-Burch, and Dolan}]{xu2015semeval}
Wei Xu, Chris Callison-Burch, and Bill Dolan. 2015.
\newblock \href {https://doi.org/10.18653/v1/S15-2001} {{S}em{E}val-2015 task
  1: Paraphrase and semantic similarity in {T}witter ({PIT})}.
\newblock In \emph{Proceedings of the 9th International Workshop on Semantic
  Evaluation ({S}em{E}val 2015)}, pages 1--11, Denver, Colorado. Association
  for Computational Linguistics.

\bibitem[{Xue et~al.(2021)Xue, Constant, Roberts, Kale, Al-Rfou, Siddhant,
  Barua, and Raffel}]{mt5}
Linting Xue, Noah Constant, Adam Roberts, Mihir Kale, Rami Al-Rfou, Aditya
  Siddhant, Aditya Barua, and Colin Raffel. 2021.
\newblock \href {http://arxiv.org/abs/2010.11934} {mt5: A massively
  multilingual pre-trained text-to-text transformer}.

\end{thebibliography}
\bibliographystyle{acl_natbib}

\appendix

\section*{Appendix}
\label{sec:appendix}

\section{Baseline Code-Mixing Generation Methods}
\subsection{Monolingual Source Paraphrasing}
\label{sec:msp}
To generate paraphrases, we use the T5\textsubscript{BASE}~\cite{t5} model finetuned on paraphrases from diverse English sources: paraphrase and semantic similarity in Twitter shared task (PIT-2015)~\cite{xu2015semeval},  {LanguageNet} (tweet)~\cite{lan2017continuously}, {Opusparcus} ~\cite{creutz2018open}  (video subtitle), and Quora question pairs (Q\&A website). \footnote{\url{https://www.quora.com/q/quoradata/First-Quora-Dataset-Release-Question-Pairs}} For all datasets excluding Quora question pairs, we keep sentence pairs with a semantic similarity score $\geq 70\%$. We merge all the datasets, split the resulting data into training, validation, and testing ($80\%$, $10\%$, and $10\%$). The T5 model is finetuned on the training split for $20$ epochs with constant learning rate of $3e^{-4}$. Given an English sentence to paraphrase, the finetuned model uses top-$p$ sampling~\cite{holtzman_iclr20} during inference to generate $10$ diverse paraphrases. We pick relevant paraphrases for a given sentence by ranking all the generated paraphrases based on the semantic similarity score with the original English sentence and discarding those paraphrases whose semantic similarity score $\geq 95\%$.

\section{Performance With mT5 On Custom Splits}
\label{sec:appd_mt5_perf}
Table~\ref{tab:cus_splits_apd} presents the performance of our proposed system on custom splits, along with best epoch for each stage in the pipeline.

\begin{table*}[t]
\footnotesize
\centering
\begin{tabular}{p{.9in}p{0.8in}p{0.8in}p{0.8in}p{0.8in}}  \hline
cs method (hyper.) & S1 epoch &  S2 epoch & Valid & Test \\ \hline
\textit{Romanization}  \\ 
IIT-B (100K) & 3 & 50 & 14.27 & 12.95 \\
IIT-B (250K) & 5 & 47 & \textbf{14.74} & \textbf{12.75} \\
IIT-B (500K) & 3 & 46 & 14.12 & 12.46 \\ 
OPUS (100K) & 3 & 43 & 14.67 & 12.62 \\
OPUS (250K) & 3 & 50 & 14.57 & 12.71 \\ \hline
\textit{Paraphrasing}  \\ 
Para (1) & 5 & 43 & 14.39 & 12.72 \\
Para (2) & 5 & 43 & 14.4 & 12.62 \\
Para (3) & 5 & 44 & \textbf{15.07} & \textbf{12.63} \\ \hline
\textit{Backtranslation} \\
Forward model & 3 & 37 & \textbf{14.07} & \textbf{12.16} \\
Backward model & 3 & 36 & \textbf{14.51} & \textbf{13.03} \\ \hline
\textit{Social media} \\
PHINC & 5 & 29 & \textbf{14.71} & \textbf{12.68} \\ \hline
\textit{CMDR (ours)}
 \\ 
CMDR-unigram & 3 & 48 & \textbf{14.6} & \textbf{12.69} \\
CMDR-bigram & 5 & 42 & 14.58 & 12.4 \\ \hline
\end{tabular}
\caption{Performance in BLEU of mT5 model finetuned using curriculum learning --- finetuning on one of the different code-mixed data generation method (max. epochs is 5) followed by finetuning on gold data (max. epochs is 50). \textbf{CMDR:} Code-Mixing from Distributed Representations refers to our proposed method. Validation performance is calculated on 541 randomly picked examples from the official training set after deduplication, while the rest of the $7,000$ examples are used for training. Test performance is calculated on the official validation set. Note that we did not study the method based on equivalence constraint theory in this experiment. For CMDR, we perform $n$-gram translation of Hindi from native to Roman script.}
\label{tab:cus_splits_apd}
\end{table*}

\end{document}